\documentclass{article}

\PassOptionsToPackage{numbers, compress}{natbib}

    \usepackage[final]{neurips_2019}

\usepackage[utf8]{inputenc} 
\usepackage[T1]{fontenc}    
\usepackage{hyperref}       
\usepackage{url}            
\usepackage{booktabs}       
\usepackage{amsfonts}       
\usepackage{nicefrac}       
\usepackage{microtype}      
\usepackage{bm}
\usepackage{amsmath}
\usepackage{graphicx}
\usepackage{multirow}
\usepackage[page]{appendix}
\usepackage{algpseudocode}
\usepackage{algorithm}

\usepackage{subcaption}
\usepackage{float}
\usepackage{placeins}

\usepackage{tikz}
\usetikzlibrary{positioning}
\usetikzlibrary{decorations.text}
\usetikzlibrary{decorations.pathmorphing}
\usetikzlibrary{shapes,snakes}

\newtheorem{theorem}{Theorem}[section]

\newtheorem{definition}[theorem]{Definition}

\graphicspath{{./images/}}

\title{Differentially Private \\ Federated Variational Inference}

\author{%
  Mrinank Sharma,$^1$\thanks{Equal contribution. Work done whilst at the University of Cambridge. Correspondence to Mrinank Sharma, \texttt{<mrinank@robots.ox.ac.uk>.}}\ \ Michael Hutchinson,$^1$\footnotemark[1] \ Siddharth Swaroop,$^2$ \\
  \bf{Antti Honkela,$^3$ Richard E. Turner$^2$}
  \\ $^1$ University of Oxford, UK. \quad
   $^2$ University of Cambridge, UK
   \\$^3$ University of Helsinki, Finland
}

\begin{document}
\maketitle
\begin{abstract}
    In many real-world applications of machine learning, data are distributed across many clients and cannot leave the devices they are stored on. Furthermore, each client's data, computational resources and communication constraints may be very different. This setting is known as federated learning, in which privacy is a key concern. Differential privacy is commonly used to provide mathematical privacy guarantees. This work, to the best of our knowledge, is the first to consider federated, differentially private, Bayesian learning. 
    We build on Partitioned Variational Inference (PVI) which was recently developed to support approximate Bayesian inference in the federated setting. We modify the client-side optimisation of PVI to provide an $(\epsilon, \delta)$-DP guarantee. We show that it is possible to learn moderately private logistic regression models in the federated setting that achieve similar performance to models trained non-privately on centralised data.
\end{abstract}

\section{Introduction}
In many real-world settings, machine learning practitioners are tasked with learning good models using data fragmented across a number of clients, such as data held on mobile phones or Internet of Things devices, or patient data held in different hospitals' databases. Each client may have large differences in the distribution of data held, network connectivity, and computational resources. This problem setting is known as federated learning setting, in which privacy is a key concern. Whilst federated learning algorithms do not directly communicate locally held data when training their models, it has been shown that modern trained machine learning models, such as neural networks, can unintentionally memorise their training examples \cite{Carlini2018}. When the training data includes highly sensitive data e.g. when performing drug sensitivity prediction with genomic data \cite{Honkela2018}, it is vital to protect the privacy of each individual who has contributed their data. Differential privacy (DP) \cite{Dwork2006} achieves this by introducing carefully calibrated noise, thus providing individuals with plausible deniability and admitting mathematical guarantees. In addition to privacy concerns, it is essential to have accurate uncertainty in predictions, particularly when they are used to make important decisions. Bayesian inference is one principled approach to reasoning under uncertainty.

%


This work provides a framework for performing federated, differentially private, Bayesian learning. 
To the knowledge of the authors, there has been very limited work in the intersection of these fields. The \emph{Federated Averaging} algorithm \cite{McMahan2016} performs local gradient descent steps to update a shared model in parallel across a number of clients, and performs an update by performing a weighted average in which clients with more datapoints are weighted more highly. This algorithm has also been adapted to be DP \cite{McMahan2018}. This algorithm does not support more challenging federated settings such asynchronous settings where the clients control when communication occurs. Previous work has also performed differentially private variational inference (an approximate Bayesian algorithm), both for non-conjugate models \cite{Jalko2016} and for conjugate-exponential family models \cite{Park2016}, but this work has not considered the federated setting. Bayesian learning on private distributed data has been performed using standard DP techniques \cite{Heikkila2017}, but this is only applicable for exponential conjugate family models. In this work we use the Partitioned Variational Inference (PVI) algorithm, a recently developed approach to variational inference that supports federated learning in conjugate and non-conjugate models, extending it to be differentially private.

\section{Background}

\subsection{Partitioned Variational Inference (PVI)}

In global Variational Inference (VI) \cite{jordan1999}, we seek to find a variational distribution which is close to the true intractable Bayesian posterior by minimising the KL divergence between the variational distribution and the true posterior, or equivalently optimising a global free energy. 

We now introduce the Partitioned Variational Inference (PVI) algorithm \cite{Bui2018}. Consider a parametric probabilistic model given by the prior $p(\bm{\theta})$ over the unknown parameters, $\bm{\theta}$, and likelihood, $p(\bm{y}| \bm{\theta})$. The data, $\bm{y}$, is spread across $M$ clients, $\lbrace \bm{y}_1, \ldots, \bm{y}_M \rbrace$ and each client communicates with a central parameter server. We use a variational distribution, $q(\bm{\theta})$, to approximate the true posterior, $p(\bm{\theta} | \bm{y})$:
\begin{align}
    q(\bm{\theta}) = p(\bm{\theta}) \prod_{m=1}^{M} t_m(\bm{\theta}) \simeq \frac{1}{\mathcal{Z}} p(\bm{\theta}) \prod_{m=1}^{M} p(\bm{y}_m | \bm{\theta}) = p(\bm{\theta} | \bm{y}).
\end{align}
Here $t_m(\bm{\theta})$ will be refined to approximate its corresponding (un-normalised) likelihood, $p(\bm{y}_m | \bm{\theta})$. The $t_m(\bm{\theta})$'s are exponential family distributions, making computation relatively simple, like in global VI. Instead of optimising the global free energy directly, a set of client-side local optimisations are performed. The $m$-th client updates as in Equation \ref{eq:pvi_optimisation}. The updated approximate likelihood $t_{m}^{\text{new}}(\bm{\bm{\theta}})$ is found by division (see Equation \ref{eq:PVI_likelihood_factor}), and the change in likelihood factor $\Delta t_{m}(\bm{\bm{\theta}})$ is communicated to the parameter server.
%
%
%
It can be shown that any fixed point of PVI is a global VI solution  \cite{Bui2018}. 
Unlike global VI, PVI maintains  an explicit representation of the contribution from each client to the approximate posterior, and refines it locally. This means it is ideally suited to the federated learning setting, particularly in inhomogeneously distributed cases.

%
    
\subsection{Differential Privacy (DP)}
\begin{definition}[$(\epsilon, \delta)$-Differential Privacy]
A randomised algorithm, $\mathcal{A}$, is said to be $(\epsilon, \delta)$ differentially private if for any possible subset of outputs, $S$, and for all datasets which differ in one entry only, $(\mathcal{D}, \mathcal{D'})$ the following inequality holds \cite{Dwork2013} :
\begin{align}
    \text{Pr}(\mathcal{A}(\mathcal{D} \in S)) \leq e^{\epsilon} \text{Pr}(\mathcal{A}(\mathcal{D}' \in S)) + \delta.
\end{align}
\end{definition}
 Additionally, modern techniques rely heavily on \emph{privacy amplification by subsampling} \cite{Li2012} i.e., randomly selecting data-points to contribute in a DP mechanism. Composition techniques enable the cumulative privacy loss to be tracked when multiple randomised algorithms are applied to a dataset. One technique for composition is known as the \emph{Moments Accountant} \cite{Abadi2016}, which provides tight bounds on the privacy cost and takes into account the privacy amplification effect of subsampling.

The DP-SGD algorithm \cite{rajkumar2012} adapts standard stochastic gradient descent by aggregating gradients with fixed $\ell_2$-norm (which may be achieved by gradient rescaling) and injecting noise. See Equation \ref{eq:DP gradients}. It can be shown that this algorithm is $(\epsilon, \delta)$-differentially private. We use a similar algorithm to optimise the PVI local objective function, using Adagrad instead of SGD.

    
\begin{algorithm}[t]
	\caption{Differentially Private Partitioned Variational Inference (DP-PVI)}
	\label{alg:DP-PVI}
	\begin{algorithmic}[1] 
	    \State \textbf{Input:} Clients $\{\bm{y}_m\}_{m=1}^M$, where $\bm{y}_m = \lbrace (\bm{x}_i, t_i) \rbrace_{i=1}^{N_m}$. Parameters: minibatch size $L$, gradient norm bound $C$, noise scale $\sigma$.
		\State Within each client, having received $q^{\text{old}}(\bm{\theta})$ from the global server, optimise:
		\begin{align}
            q^{\text{new}}(\bm{\theta}) = \underset{q(\bm{\theta}) \in \mathcal{Q}}{\text{arg min}}\quad \mathcal{KL}\Big(q(\bm{\theta}) || \frac{1}{\mathcal{Z}'} \frac{q^{\text{old}}(\bm{\theta})}{t_{m}^{\text{old}}(\bm{\theta})} p(\bm{y}_m | \bm{\theta}) \Big).
            \label{eq:pvi_optimisation}
        \end{align}
		\State This optimisation is done via Adagrad. At each iteration $t$, use the Gaussian Mechanism on the minibatch gradient, subsampling a minibatch of size $L$ (denoted as $\mathcal{L}$):
		\begin{align}
		    \bm{\tilde{g}}_t = \frac{1}{L} \left[ \sum_{i \in \mathcal{L}}{\frac{\bm{g}(\bm{x}_i)}{\text{max}\left(1, \frac{\lVert\bm{g}(\bm{x}_i)\rVert_2}{C} \right)}} + \mathcal{N}(0, \sigma^2 C^2 \bm{I}) \right].
		    \label{eq:DP gradients}
		\end{align}
		\State After optimisation, communicate to the global server:
		\begin{align}
            \Delta t_{m}(\bm{\theta}) = \frac{t_{m}^{\text{new}}(\bm{\theta})}{t_{m}^{\text{old}}(\bm{\theta})} = \frac{q^{\text{new}}(\bm{\theta})}{q^{\text{old}}(\bm{\theta})}.
            \label{eq:PVI_likelihood_factor}
        \end{align}
        \State The global server updates $q(\bm{\theta}) \leftarrow q(\bm{\theta}) \Delta t_{m}(\bm{\theta})$.
	\end{algorithmic}
\end{algorithm}

\section{Experiments}
\textbf{Method.} We consider binary classification on the standard UCI Adult dataset \cite{Dua:2019} (also considered in \citet{Jalko2016}) with an $80\%:20\%$ train:test split, predicting whether household income exceeds $\$50,000$ based on sensitive factors such as occupation, relationship status and native country, making individual (datapoint) privacy important. We specifically focus on unbalanced distributions of data across clients. In order to achieve this, we split the clients into one group of smaller clients and another group of larger clients. We change the proportion of datapoints in each group, and the class balance of each group (see Appendix \ref{appendix:data_dist}). We also focus on the asynchronous setting, where the parameter server is not in control of when a client communicates an update. We assume each client has similar compute power, and thus has update rate inversely proportional to the number of data points held by that client. In this case, we expect that PVI will outperform global VI, and we show this in Appendix \ref{appendix:global_vs_pvi} for a wide range of dataset distributions.

We perform mean-field PVI using a logistic regression model with a Gaussian prior and variational distribution (See Appendix~\ref{appendix:model_definition} for the full model definition). We use the Gaussian mechanism to provide a private gradient estimate and apply Adagrad to perform the client-side PVI optimisation
(Equation~\ref{eq:pvi_optimisation}), providing communicated client updates with a DP guarantee (See Appendix~\ref{appendix:threat_model}) using the Moments Accountant to compose privacy guarantees \cite{Abadi2016}. At each client, we choose $\delta$ to be an order of magnitude lower than the number of locally held data points. The maximum privacy budget ($\epsilon_\text{max}$) is specified for each client, and a client no longer communicates an update if this budget is exceeded. We compare DP-PVI with non-private global VI and PVI using gradient ascent.


\begin{figure}[h]
    \centering
    
    \begin{subfigure}[b]{0.48\textwidth}
        \includegraphics[width=\textwidth]{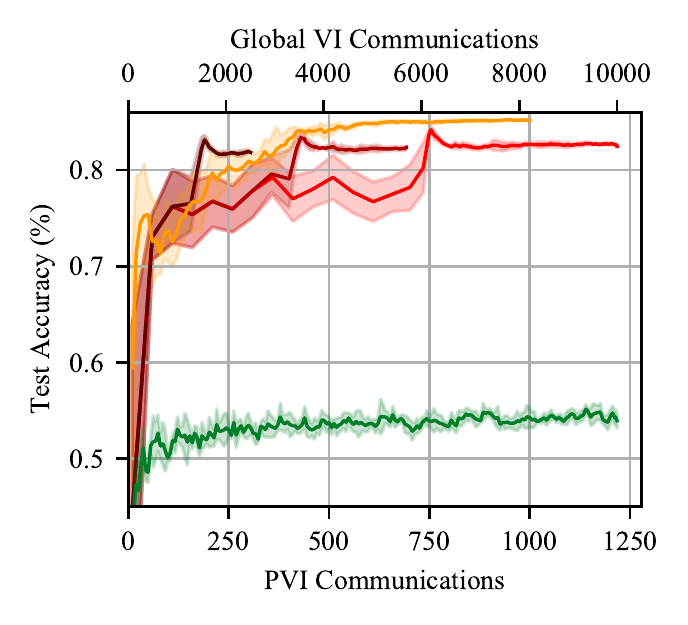}
    \end{subfigure}
    \begin{subfigure}[b]{0.48\textwidth}
        \includegraphics[width=\textwidth]{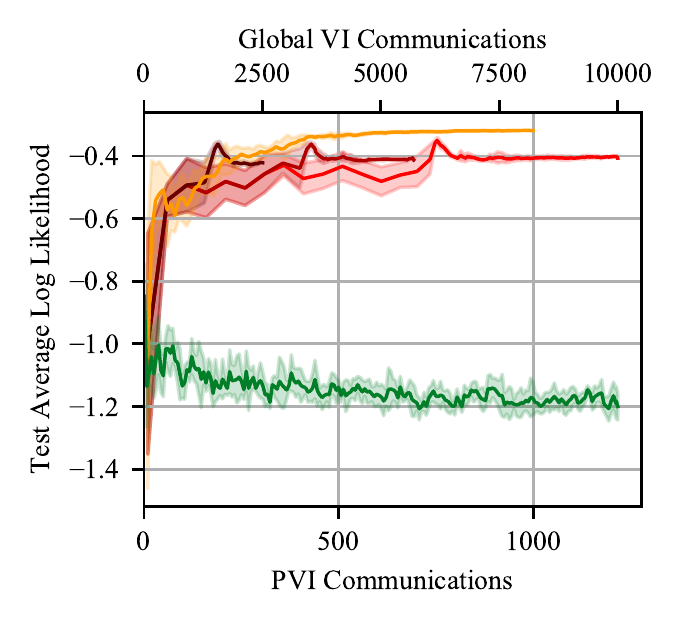}
    \end{subfigure}
    \\
    \vspace{-1em}
    \begin{subfigure}[b]{0.48\textwidth}
        \includegraphics[width=\textwidth]{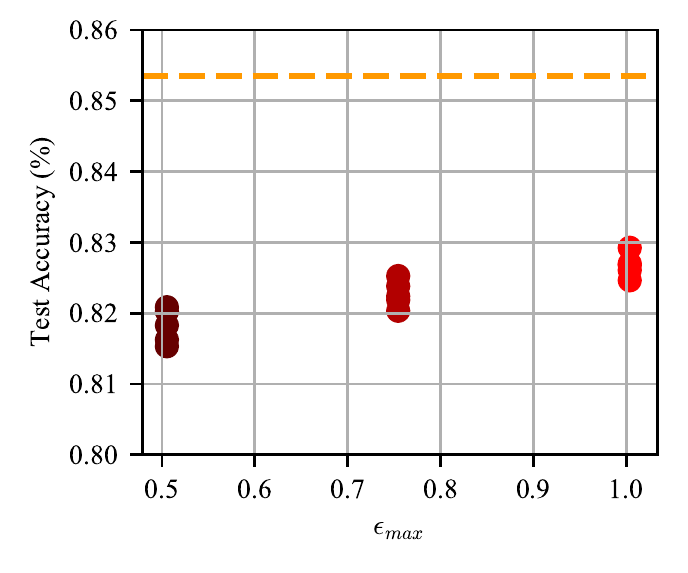}
    \end{subfigure}
    \begin{subfigure}[b]{0.48\textwidth}
        \includegraphics[width=\textwidth]{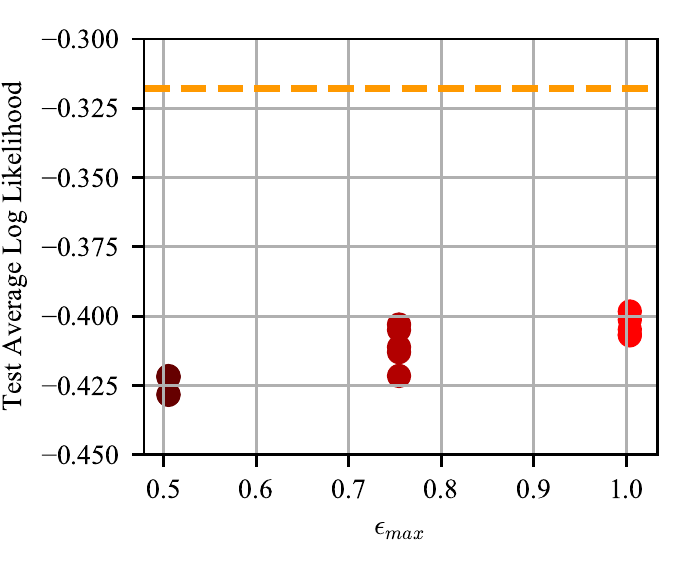}
    \end{subfigure}
    \\
    \vspace{-0.5em}
    \includegraphics[width=0.8\textwidth]{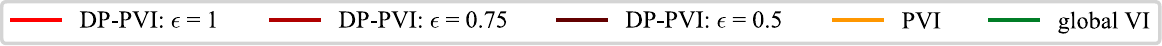}
    
    \caption{Dataset Distribution C (inhomogenous) Results. Left: test set accuracy. Right: test set average log likelihood. Upper: performance as a function of parameter communications. Note that the $x$-axis scale is different for global VI and PVI. Lower: privacy-utility tradeoff. PVI (and DP-PVI) performs significantly better than global VI in terms of test set accuracy and performance. DP-PVI with moderate privacy guarantees achieves performance similar to non-private PVI. Increasing $\epsilon_\text{max}$ improves performance as each client is able to participate in additional communication rounds. See Appendix~\ref{app:additional_results} for results obtained on alternative dataset distributions. \vspace{-1em}}
    \label{fig:dataset_c_results}
\end{figure}

\textbf{Results.} Fig.~\ref{fig:dataset_c_results} shows results for Dataset Distribution C (see Appendices \ref{appendix:data_dist} and \ref{appendix:hyperparam_settings} for the data set distribution settings and hyperparameter settings respectively)\footnote{To reproduce the experiments in this paper, please see \url{https://github.com/MrinankSharma/DP-PVI}}. Each algorithm is run on 5 randoms seeds and the mean and standard deviation of results reported. As expected, we find that non-private PVI outperforms non-private global VI when comparing both held-out log likelihood and test set accuracy as a function of communication rounds. Not only is each communication between a server and client more meaningful in PVI, since it has been performed with several update steps, but also the factorisation of the variational distribution employed in PVI means that smaller clients updating more frequently should not have an adverse effect. The average gradient size is independent of the number of data points held by that client for global VI. 

We find that the DP-PVI algorithm is able to achieve close to non-private performance for a moderate privacy guarantee. Furthermore, if we reduce the privacy guarantee by increasing $\epsilon_\text{max}$ for each client, the performance of the algorithm improves slightly.
We remark that the training curves for the DP-PVI algorithm show  an increase in performance followed by a small but consistent decrease in performance near the end of training for all values of $\epsilon_\text{max}$. We believe that the reason for this behaviour is as follows:
as client $m$ updates with probability inversely proportional to $N_m$, clients with small $N_m$ update more frequently, meaning they expend their privacy budget earlier on in training. The decrease in performance is due to clients with large $N_m$ repeatedly updating at the end of training, and since these clients are not representative of the over all data distribution, we see a corresponding decrease in performance. 

The results for alternative data distributions are in Appendix \ref{app:additional_results}. The DP-PVI algorithm attains a significantly better accuracy and held out log-likelihood than the \emph{non-private} global VI approach, except for the homogeneous case where the performance is similar. In all cases DP-PVI requires substantially fewer communication rounds. Moreover, global VI on homogeneous data is equivalent to centralised VI, representing  gold-standard performance. In this context, the worst case test set accuracy and average log likelihood of DP-PVI is $81.83\%$ and $-0.4218$ (Dataset C, $\epsilon_\text{max} = 0.5)$ which is similar to centralised VI's performance of $85.21\%$ and $ -0.3184$. 


\section{Conclusions}

In this work we introduce a first of its kind method for performing differentially private, federated, Bayesian learning by augmenting the client side optimisations in PVI to yield an $(\epsilon, \delta)$-DP guarantee. We find that this technique achieves similar performance to the current non-private state-of-the-art method (PVI), while providing moderate privacy guarantees. 

The approach taken here is appropriate when each individual contributes only one datapoint in total, which is the precisely the scenario for the UCI Adult data set. However, \emph{client level differential privacy} may be more natural when the data from each client corresponds to one individual only, and this would consider adjacent data sets to be those which differ in the data of one client only. We believe that this will be a fruitful direction for future work. 


\bibliography{references}
\newpage

\appendix
\section{Threat Model}
\label{appendix:threat_model}
\tikzstyle{client} = [fill=blue!20, rounded corners, 
text centered, circle, minimum size = 6em, text width=6em]
\tikzstyle{server} = [draw, fill=red!30, rounded corners, 
text centered, rectangle, minimum size = 6em, text width = 10em]
\tikzstyle{adv} = [draw, fill=red!30, align=center,
text centered, ellipse, minimum size = 4em, text width = 4em]
\tikzstyle{info} = [draw, fill=orange, fill opacity=0.2, rounded corners, 
text centered, rectangle, minimum height = 8em, align=left,
text width = 8em]

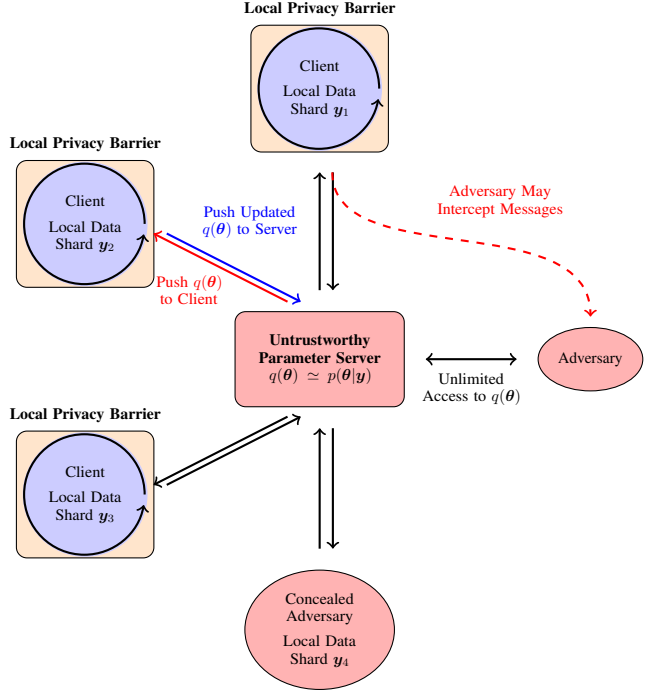
\begin{figure}[H]
    \centering
    \begin{tikzpicture}[scale=0.6, every node/.style={transform shape}]
\node [server] (srv) at (0:0) {\textbf{Untrustworthy Parameter Server} \\ $q(\bm{\theta}) \simeq p(\bm{\theta}| \bm{y})$};
\node[info] (info1) at (150:6) {};
\node [client] (cl1) at (150:6) {Client \\\vspace{0.5em} Local Data Shard $\bm{y}_2$};
\node [text centered, rectangle, minimum size = 4em, text width = 12em, yshift=10pt] (textNode) at (info1.north) {\textbf{Local Privacy Barrier}};
\node[info] (info2) at (90:6) {};
\node [client] (cl2) at (90:6) {Client \\\vspace{0.5em} Local Data Shard $\bm{y}_1$};;
\node [text centered, rectangle, minimum size = 4em, text width = 12em, yshift=10pt] (textNode) at (info2.north) {\textbf{Local Privacy Barrier}};
\node[info] (info3) at (210:6) {};
\node [text centered, rectangle, minimum size = 4em, text width = 12em, yshift=10pt] (textNode) at (info3.north) {\textbf{Local Privacy Barrier}};
\node [client] (cl3) at (210:6) {Client \\\vspace{0.5em} Local Data Shard $\bm{y}_3$};
\node [client, fill=red!30, ellipse, draw] (cl4) at (270:6) {Concealed Adversary \\\vspace{0.5em} Local Data Shard $\bm{y}_4$};;
\node[adv] (adv) at (0:6) {Adversary};
\draw [thick, ->] ([shift={(-0.2em,0)}]cl1.east) arc (0:350:3.8em);
\draw [thick, ->] ([shift={(-0.2em,0)}]cl2.east) arc (0:350:3.8em);
\draw [thick, ->] ([shift={(-0.2em,0)}]cl3.east) arc (0:350:3.8em);
\draw [thick,transform canvas={xshift=-5pt},  ->, shorten >= 8pt, shorten <= 8pt, color=red] (srv.north) to node [text width=2.5cm,midway,below=0.1em, scale = 1,xshift=-2em, align=center] {Push $q(\bm{\theta})$ to Client} (cl1.east);
\draw [thick,->, shorten >= 8pt, shorten <= 8pt, color=blue] (cl1.east) to node [text width=2.5cm,midway,above=1.5em, scale = 1, xshift=10pt, align=center] {Push Updated  $q(\bm{\theta})$ to Server}  (srv.north);

\draw [thick, transform canvas={xshift=5pt},  ->, shorten >= 8pt, shorten <= 8pt] (cl2.south) to (srv.north);
\draw [thick, ->, shorten >= 8pt, shorten <= 8pt] (srv.north) to (cl2.south);

\draw [thick, transform canvas={xshift=-5pt},  ->, shorten >= 8pt, shorten <= 8pt] (srv.south) to (cl3.east);
\draw [thick, ->, shorten >= 8pt, shorten <= 8pt] (cl3.east) to (srv.south);

\draw [thick, transform canvas={xshift=5pt},  ->, shorten >= 8pt, shorten <= 8pt] (srv.south) to (cl4.north);
\draw [thick,->, shorten >= 8pt, shorten <= 8pt] (cl4.north) to (srv.south);

\draw [thick, <->, shorten >= 8pt, shorten <= 8pt] (adv.west) to node [text width=2.5cm,midway,below=0.5em, scale = 1, align=center] {Unlimited Access to $q(\bm{\theta})$} (srv.east);

\draw[thick, ->,  shorten >= 4pt, shorten <= 8pt, red, dashed] (cl2.south) ++(8pt, 0) to [out=270, in=90] (adv.north);

\node [text centered, rectangle, minimum size = 4em, text width = 10em, color=red] (textNode) at (4, 3.5) {Adversary May Intercept Messages};
\end{tikzpicture} 
    \caption{Threat model assumed by the DP-PVI algorithm.}
    \label{fig:threat_model}
\end{figure}

Fig. \ref{fig:threat_model} shows the threat model implied by the DP-PVI algorithm. Since the updates produced by each client are individually protected by an $(\epsilon, \delta)$ guarantee with an appropriate value of $\delta$, the algorithm does not require secure transmission protocols, nor does it assume that each client contributing data is trustworthy or that the central parameter server is genuine. This is a useful situation, and each client is in direct control of the privacy of their own data.

\section{Model Definition}
\label{appendix:model_definition}
We consider the problem of binary classification. Client $m$ holds data $\bm{y}_m = \lbrace (\bm{x}_i, t_i) \rbrace_{i=1}^{N_m}$ where $\bm{x}_m \in \mathbb{R}^d$ and $t_i \in \lbrace -1, +1\rbrace$. The model is defined by:
\begin{align}
    P(t_i | \bm{x}_i, \bm{w}) = \sigma (t_i \bm{w}^\text{T} \bm{x}_i),
\end{align}
where $\sigma(\cdot)$ is the logistic function i.e.,
\begin{align}
    \sigma(z) = \frac{1}{1 + \exp(-z)}
\end{align}
A Gaussian prior is placed on the weights, $\bm{w}$.
\begin{align}
    P(\bm{w}) = \mathcal{N}(\bm{w}| \bm{\mu}_{w}, \bm{\Sigma}_w)
\end{align}
The variational distribution applied is a mean-field multivariate Gaussian:
\begin{align}
    q(\bm{w}) = \mathcal{N}(\bm{w}| \bm{\mu}_q, \bm{\Sigma}_q),
\end{align}
where $\bm{\Sigma}_q$ is a diagonal matrix. We approximate the posterior predictive distribution by using the \emph{Probit approximation} \cite{murphy2012machine}:
\begin{align}
    p(t^* = 1| \bm{x}^*, \bm{y}) &= \int p(t^* = 1| \bm{x}^*, \bm{w}) p(\bm{w}| \bm{y})\ d\bm{w} \\
    &\simeq \sigma \left( \frac{\bm{\mu_q}^T \bm{x}^*}{\sqrt{1 + \pi \bm{x}^{*\text{T}} \bm{\Sigma}_q \bm{x}^*/8 }} \right)
\end{align}
This approximation is directly used to compute the likelihood of each datapoint. Accuracy is computed by classifying all datapoints with $p(t^* = 1| \bm{x}^*, \bm{y}) > 0.5$ as having label $1$.

\section{Dataset Distribution Scheme}
\label{appendix:data_dist}
The UCI Adult dataset is distributed across $M=10$ clients in this work, and the degree of inhomogeneity is controlled by parameters $\kappa$, which controls the client class imbalance, and $\rho$, which controls the client size imbalance. Recall that the task considered is binary classification, and denote the fraction of the \emph{majority} class in the data (which will be distributed) as $\lambda$. For the UCI Adult dataset, $\lambda = 0.76$. We used the following procedure to distribute the data across the clients.
\begin{enumerate}
    \item Let half of the clients be `small' clients and the remaining half be `large' clients.
    \item Use $\rho$ to determine target client sizes.
    \begin{align}
        N_\textbf{small} = \Big\lfloor \frac{N_\text{total}}{M} ( 1 - \rho) \Big\rfloor. \hspace{1cm} N_\textbf{large} = \Big\lfloor \frac{N_\text{total}}{M} ( 1 + \rho)\Big\rfloor.
    \end{align}
    $\lfloor \cdot \rfloor$ denotes the floor operation. Thus $\rho \in (0, 1)$ where $\rho = 0$ means that the small and large clients are of the same size, and $\rho = 1$ gives small clients having no data. 
    \item Use $\kappa$ to determine the target majority class fraction on each client type. 
    \begin{align}
        \lambda_\text{small}^{\text{target}} &= \lambda + (1 - \lambda)\cdot \kappa. 
    \end{align}
    Since $\lambda$ refers to the fraction of the majority class, $\kappa > 1$ decreases the number of minority class labels on the small clients and vice versa. Note that $\kappa$ may be negative, and its domain depends on $\lambda$.
\end{enumerate}
Note that not all combinations of $\kappa$ and $\rho$ are valid; for example, the data-set may not contain sufficient minority class examples.

\begin{table}[H]
\center
\begin{tabular}{ccccccccc}
\multirow{2}{*}{Identifier} & \multirow{2}{*}{$\rho$} & \multirow{2}{*}{$\kappa$} & \multicolumn{3}{|c|}{Small Clients} & \multicolumn{3}{c}{Large Clients} \\
 &  &  & \multicolumn{1}{|c}{$N_m$} & $\text{Proportion +ve}$ & \multicolumn{1}{c|}{$\delta$} & $N_m$ & $\text{Proportion +ve}$ & \multicolumn{1}{c}{$\delta$} \\ \hline
A & $0$ & $0$ & $3907$ & $\sim24\%$ & $10^{-4}$ & $3907$ & $\sim 24\%$ & $10^{-4}$ \\
B & $0.9$ & $0.95$ & $390$ & $\sim 1.3\%$ & $10^{-3}$ & $7424$ & $\sim 25\%$ & $10^{-4}$ \\
C & $0.7$ & $-3$ & $1172$ & $\sim 96\%$ & $10^{-4}$ & $6642$ & $\sim 11\%$ & $10^{-4}$ \\ 
D & $0.6$ & $-1.5$ & $1562$ & $\sim60\%$ & $10^{-4}$ & $6251$ & $\sim 15\%$ & $10^{-4}$ \\
\end{tabular}

\caption{Dataset distributions generated by controlling parameters $\rho$ and $\kappa$, and resulting values of $\delta$ applied. }
\end{table}

\section{Hyperparameter Settings}
\label{appendix:hyperparam_settings}

\subsection{PVI Hyperparameter settings}
\begin{table}[H]
    \centering
    \begin{tabular}{llr}
        Hyperparameter & Description & Value \\ \hline
         $L$ & Minibatch size & 100 \\
         Learning rate & Local likelihood optimiser initial learning rate & 2.0 \\
         $N_{\text{steps}}$ & Number of local optimisation steps per PVI update & 25 \\
         Damping factor & PVI damping factor to apply to local likelihood updates & 0.1
    \end{tabular}
    \vspace{1em}
    \caption{Hyperparameter settings used for the PVI algorithm.}
\end{table}

\subsection{DP-PVI Hyperparameter settings}
\begin{table}[H]
    \centering
    \begin{tabular}{llr}
         Hyperparameter & Description & Value \\ \hline
         $C$ & $\ell_2$ clipping bound & 75 \\
         $\sigma$ & Effective noise scale & 5 \\
         $q$ & Minibatch sample proportion & 0.02 \\
         Learning rate & Local likelihood optimiser initial learning rate & 0.5 \\
         $N_{\text{steps}}$ & Number of local optimisation steps per PVI update & 25 \\
         Damping factor & PVI damping factor to apply to local likelihood updates & 0.1
    \end{tabular}
    \vspace{1em}
    \caption{Hyperparameter settings used for the DP-PVI algorithm.}
\end{table}

\subsection{Global VI Hyperparameter settings}
\begin{table}[H]
    \centering
    \begin{tabular}{llr}
        Hyperparameter & Description & Value \\ \hline
         $L$ & Minibatch size & 100 \\
         Learning rate & Local likelihood optimiser initial learning rate & 0.05 \\
    \end{tabular}
    \vspace{1em}
\caption{Hyperparameter settings used for the global VI algorithm.}
\end{table}

\newpage

\section{Additional Results}
\label{app:additional_results}

\subsection{Comparison between Global VI and PVI}
\label{appendix:global_vs_pvi}
\begin{figure}[H]
    \centering
    \includegraphics[width=\textwidth]{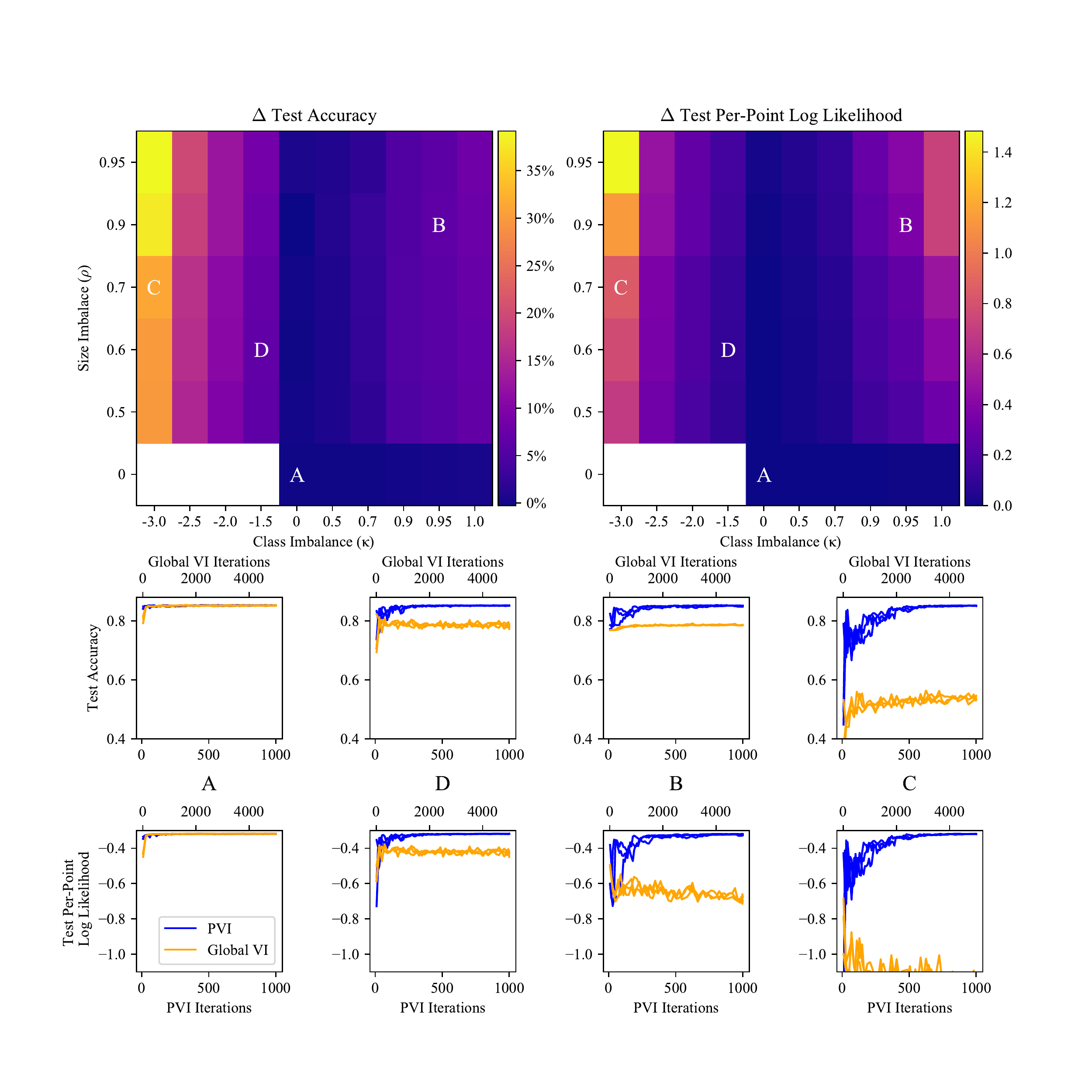}
    \caption{Upper: mean performance difference, averaged across five random seeds across the final ten iterations, between the PVI and global VI algorithms for both test accuracy and held out log likelihood for different values of $\kappa$ and $\rho$, as defined in Appendix~\ref{appendix:data_dist}. Blank cells indicate invalid settings. Lower: representative training curves for chosen values of $\kappa$ and $\rho$.}
    \label{fig:my_label}
\end{figure}

\subsection{Tabulated results}

\begin{table}[H]
\resizebox{\textwidth}{!}{%
\begin{tabular}{|l|c|c|c|c|c|c|}
\hline
\multirow{2}{*}{Dataset}       & \multicolumn{2}{l|}{A}                                                                                                            & \multicolumn{2}{l|}{B}                                                                                                            & \multicolumn{2}{l|}{C}                                                                                                            \\ \cline{2-7} 
                               & \begin{tabular}[c]{@{}l@{}}Test \\ Accuracy (\%)\end{tabular} & \begin{tabular}[c]{@{}l@{}}Average Log \\ Likelihood\end{tabular} & \begin{tabular}[c]{@{}l@{}}Test \\ Accuracy (\%)\end{tabular} & \begin{tabular}[c]{@{}l@{}}Average Log \\ Likelihood\end{tabular} & \begin{tabular}[c]{@{}l@{}}Test \\ Accuracy (\%)\end{tabular} & \begin{tabular}[c]{@{}l@{}}Average Log \\ Likelihood\end{tabular} \\ \hline
PVI                            & $ 85.23 \pm 0.05$                                             & $-0.3181 \pm 0.0003$                                              & $ 85.15 \pm 0.15$                                             & $-0.3216 \pm 0.0030$                                              & $ 85.13 \pm 0.05$                                             & $-0.3193 \pm 0.0007$                                              \\
Global VI                      & $ 85.21 \pm 0.08$                                             & $-0.3184 \pm 0.0002$                                              & $ 78.76 \pm 0.13$                                             & $-0.7006 \pm 0.0105$                                              & $ 53.96 \pm 0.78$                                             & $-1.1997 \pm 0.0438$                                              \\
DP-PVI ($\epsilon_\text{max}=0.50$)       & $ 84.57 \pm 0.00$                                             & $-0.3439 \pm 0.0000$                                              & $ 84.43 \pm 0.00$                                             & $-0.3379 \pm 0.0000$                                              & $ 81.83 \pm 0.00$                                             & $-0.4218 \pm 0.0000$                                              \\
DP-PVI ($\epsilon_\text{max}=0.75$)       & $ 84.78 \pm 0.00$                                             & $-0.3372 \pm 0.0000$                                              & $ 84.87 \pm 0.00$                                             & $-0.3332 \pm 0.0000$                                              & $ 82.38 \pm 0.00$                                             & $-0.4130 \pm 0.0000$                                              \\
DP-PVI ($\epsilon_\text{max}=1.00$)       & $ 85.02 \pm 0.15$                                             & $-0.3332 \pm 0.0018$                                              & $ 84.94 \pm 0.14$                                             & $-0.3323 \pm 0.0017$                                              & $ 82.46 \pm 0.00$                                             & $-0.4070 \pm 0.0000$                                              \\ \hline
\end{tabular}%
}
\vspace{1em}
\caption{Test accuracy and average log likelihood across different dataset distributions for the PVI, global VI and DP-PVI algorithms, considering different $\epsilon_\text{max}$. }
\end{table}

\newpage
\FloatBarrier
\subsection{Distribution A (Homogenous)}
\begin{figure}[h]
    \centering
    \begin{subfigure}{0.48\textwidth}
        \includegraphics[width=\textwidth]{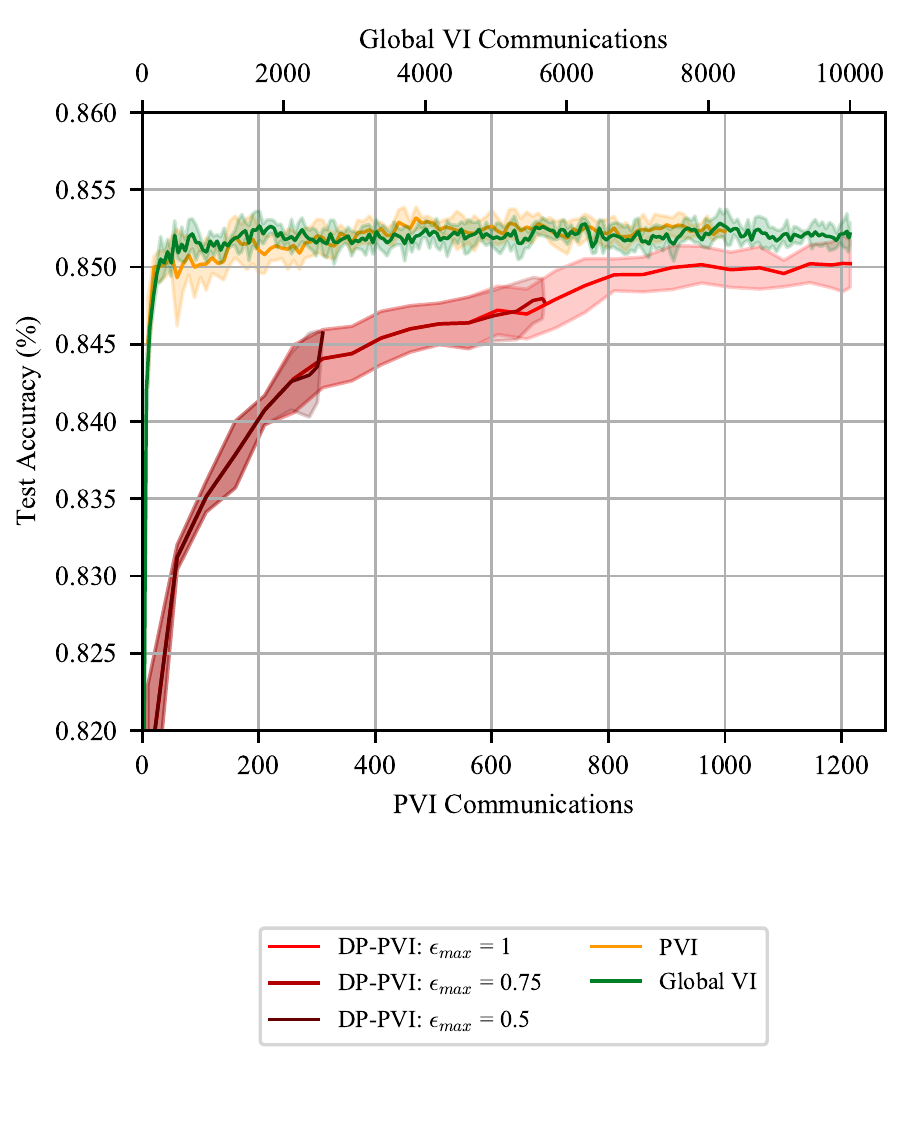}
    \end{subfigure}
    \begin{subfigure}{0.48\textwidth}
        \includegraphics[width=\textwidth]{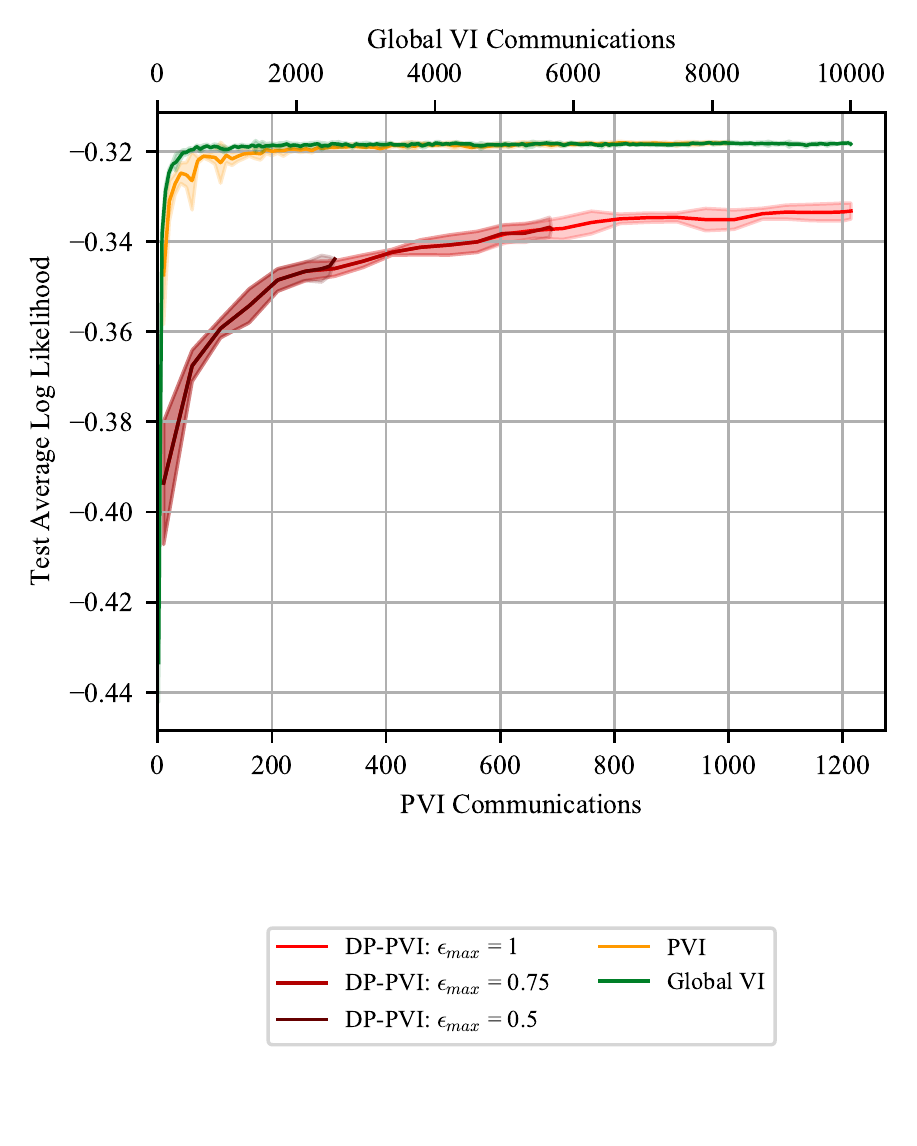}
    \end{subfigure}
    \\
    \begin{subfigure}{0.48\textwidth}
        \includegraphics[width=\textwidth]{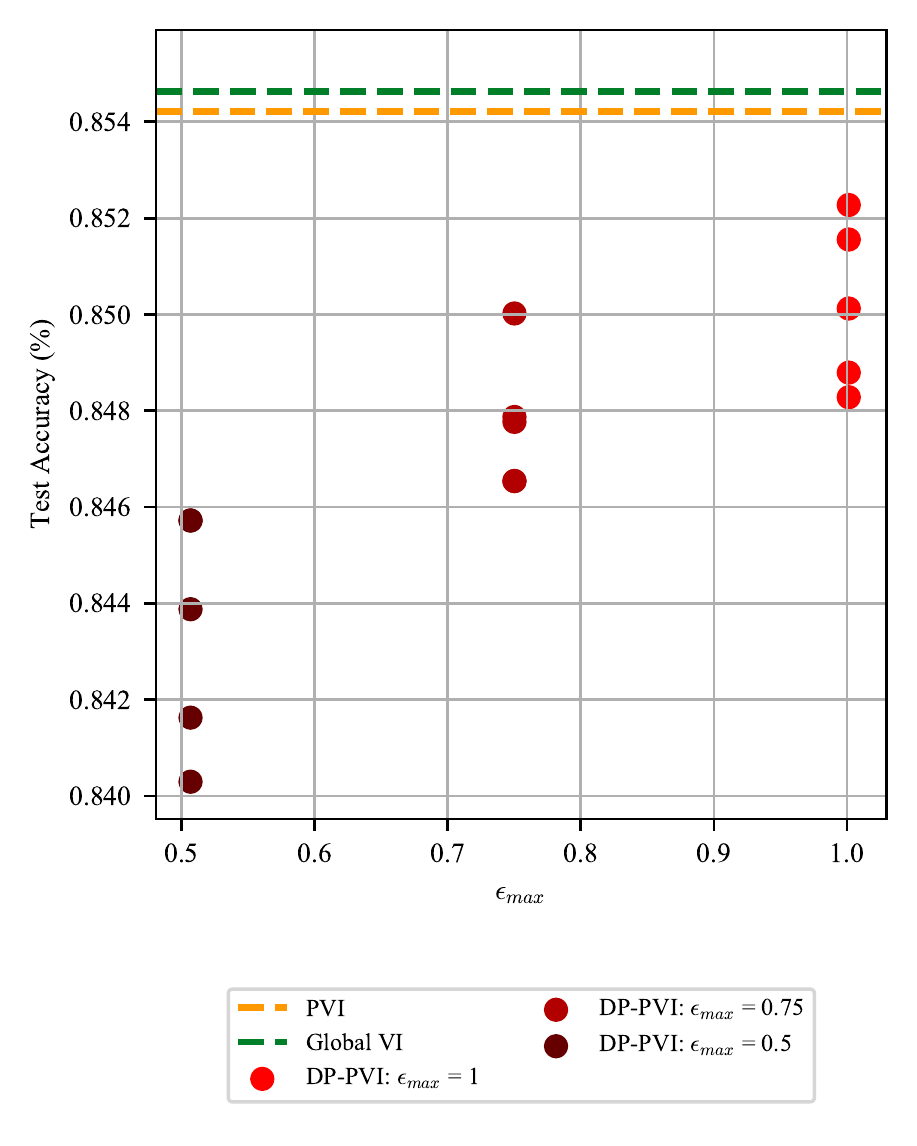}
    \end{subfigure}
    \begin{subfigure}{0.48\textwidth}
        \includegraphics[width=\textwidth]{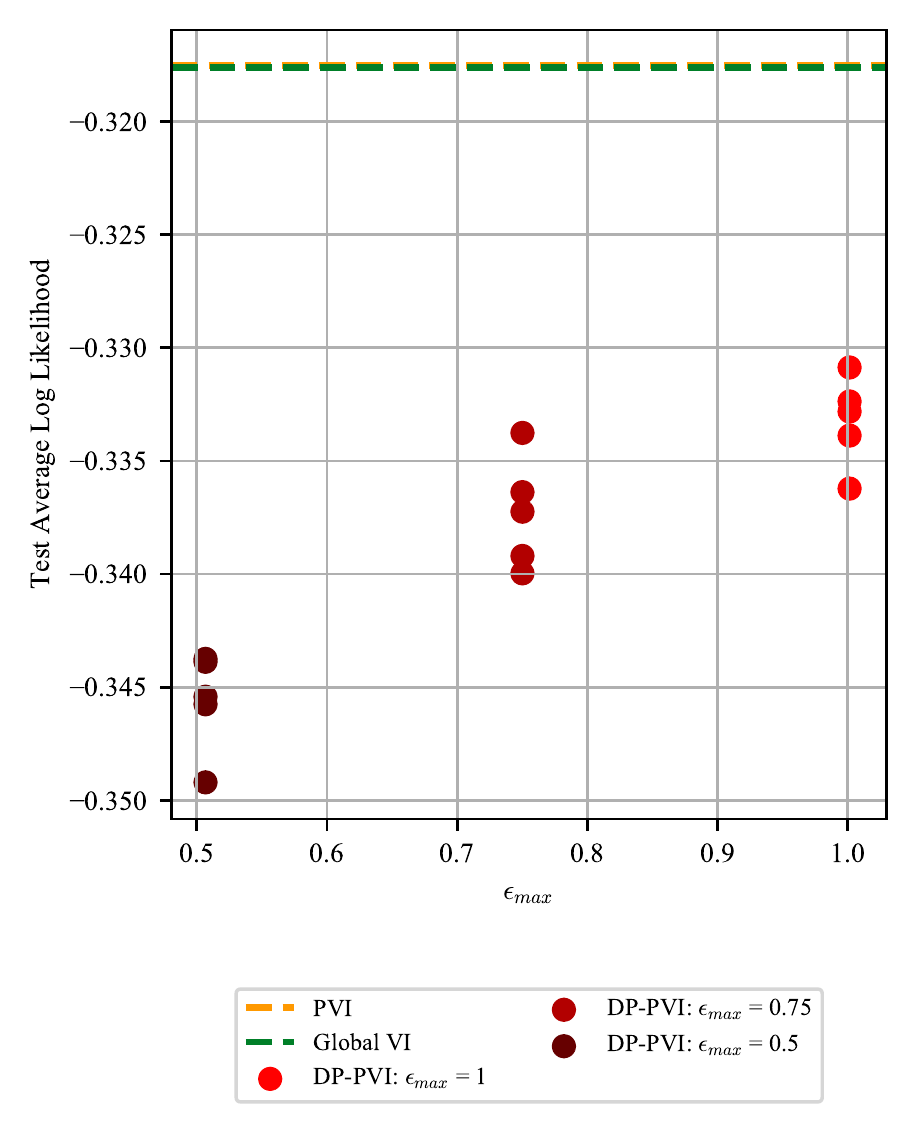}
    \end{subfigure}
    
     \caption{Dataset A (homogenous) Results. Left: test set accuracy. Right: test set average log likelihood. Upper: performance as a function of parameter communications. Note that the $x$-axis scale is different for global VI and PVI. Lower: privacy-utility tradeoff. PVI and Global VI perform similarly on homogenously distributed data. DP-PVI reaches marginly lower performance in this case. Increasing $\epsilon_\text{max}$ improves performance as each client is able to participate in additional communication rounds.}
\end{figure}

\newpage
\FloatBarrier
\subsection{Distribution B (Inhomogenous)}
\begin{figure}[h]
    \centering
    \begin{subfigure}{0.48\textwidth}
        \includegraphics[width=\textwidth]{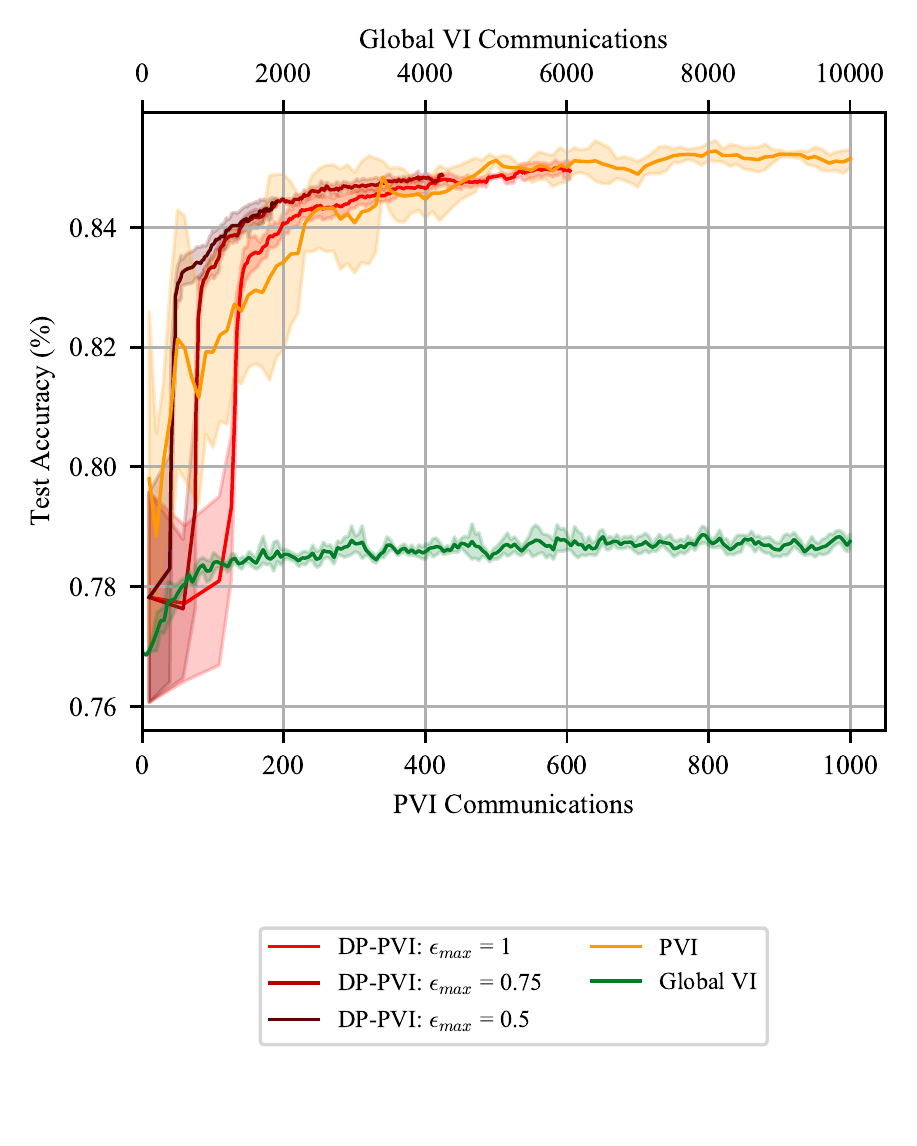}
    \end{subfigure}
    \begin{subfigure}{0.48\textwidth}
        \includegraphics[width=\textwidth]{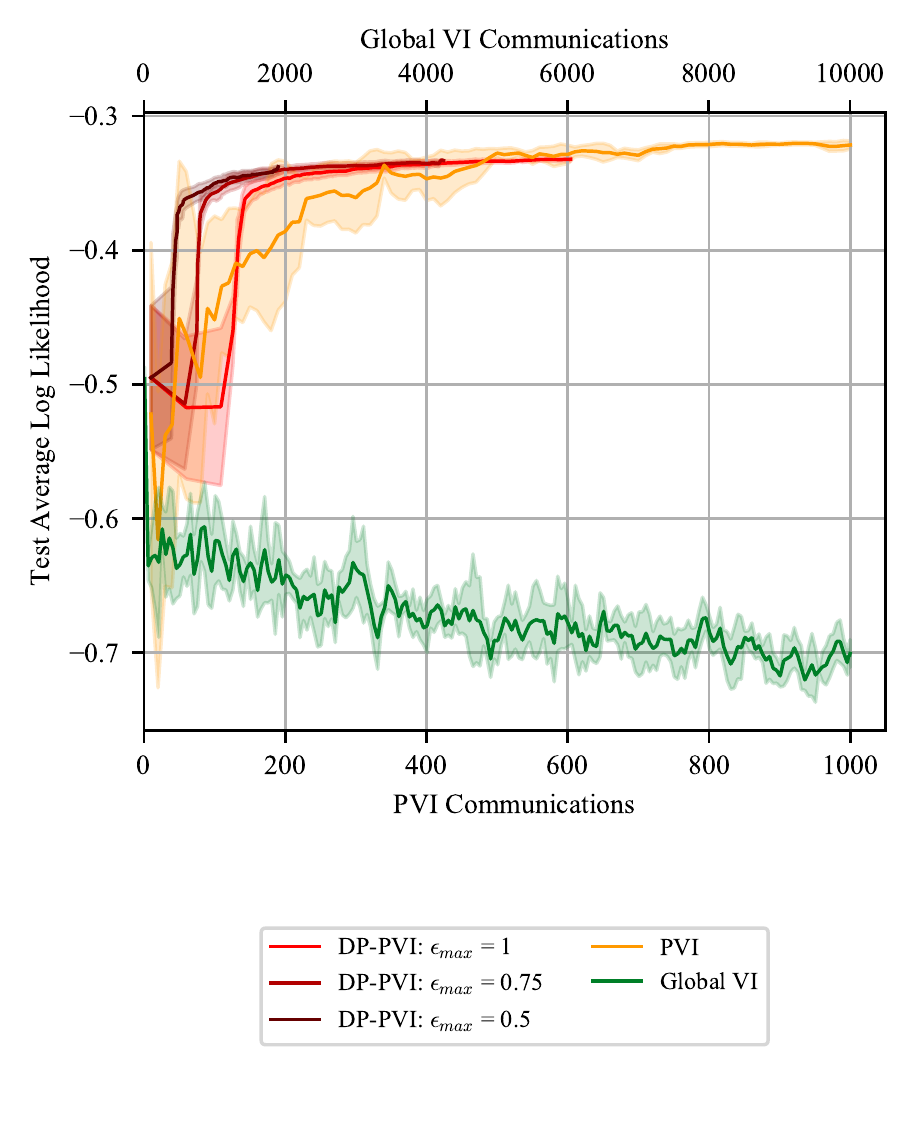}
    \end{subfigure}
    \\
    \begin{subfigure}{0.48\textwidth}
        \includegraphics[width=\textwidth]{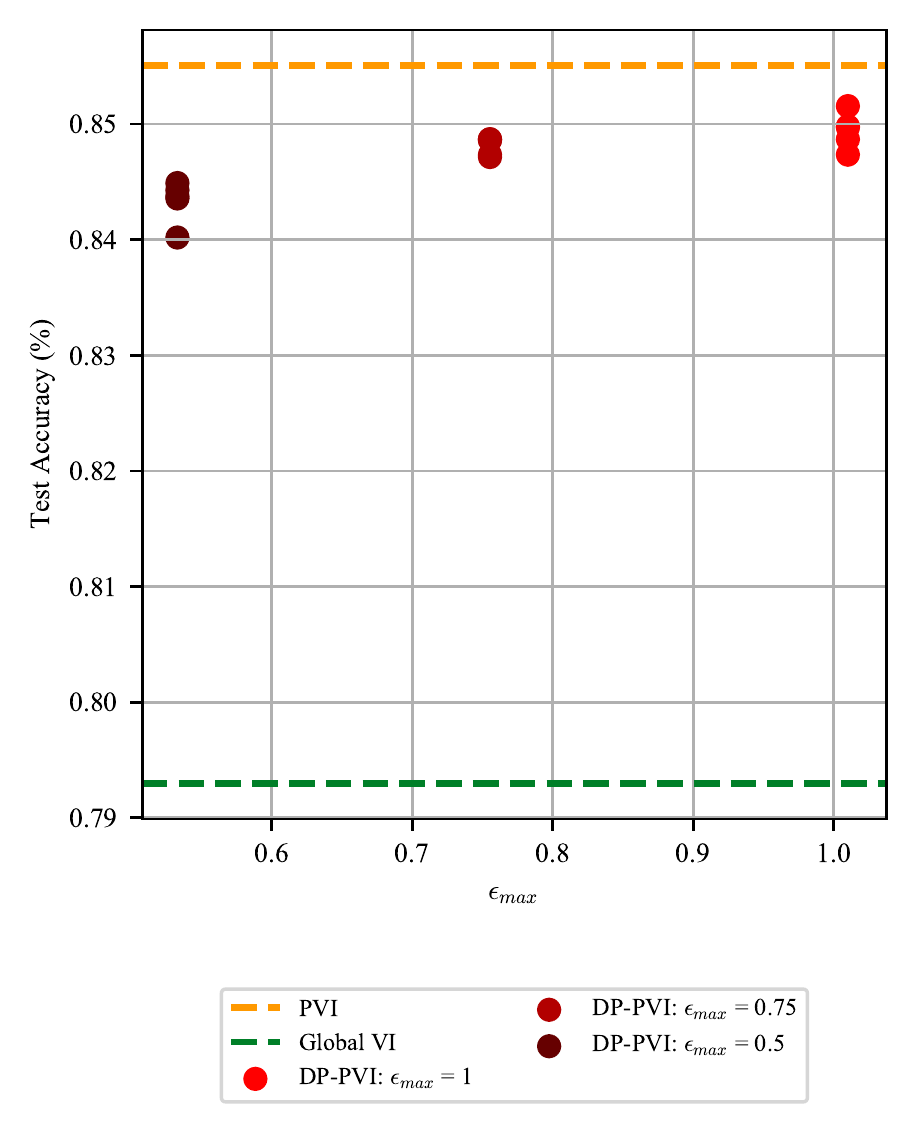}
    \end{subfigure}
    \begin{subfigure}{0.48\textwidth}
        \includegraphics[width=\textwidth]{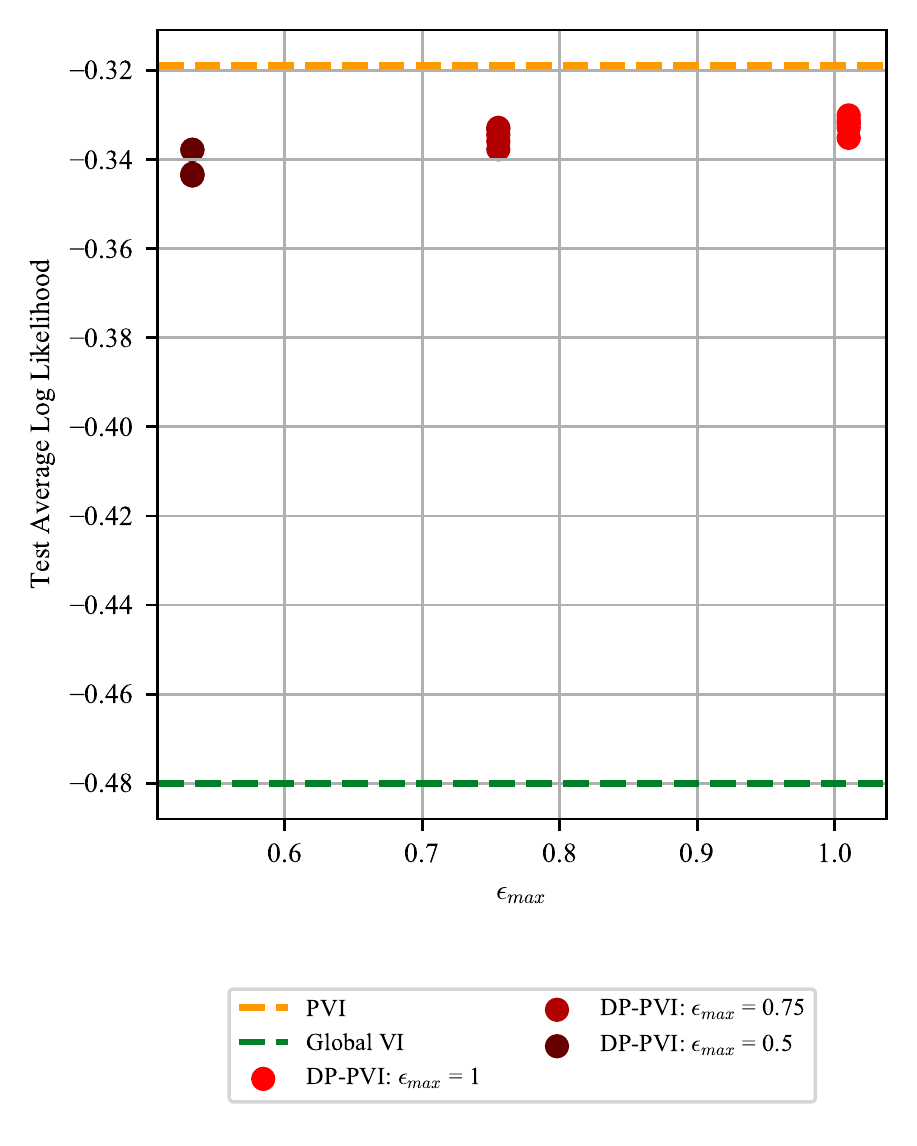}
    \end{subfigure}
    \caption{Dataset B (inhomogenous) Results. Left: test set accuracy. Right: test set average log likelihood. Upper: performance as a function of parameter communications. Note that the $x$-axis scale is different for global VI and PVI. Lower: privacy-utility tradeoff. PVI (and DP-PVI) performs significantly better than global VI in terms of test set accuracy and performance. DP-PVI with moderate privacy guarantees achieves performance similar to non-private PVI. Increasing $\epsilon_\text{max}$ improves performance as each client is able to participate in additional communication rounds.}
\end{figure}
\end{document}